\documentclass{article}

\usepackage{PRIMEarxiv}

\usepackage[utf8]{inputenc} 
\usepackage[T1]{fontenc}    
\usepackage{hyperref}       
\usepackage{url}            
\usepackage{booktabs}       
\usepackage{amsfonts}       
\usepackage{nicefrac}       
\usepackage{microtype}      
\usepackage{lipsum}
\usepackage{fancyhdr}       
\usepackage{graphicx}       
\graphicspath{{media/}}     
\usepackage{float}
\usepackage{hanging}

\title{Game of Intelligent Life
\thanks{\textit{Full Demo Visuals}: 
{https://chaytanc.github.io/gil} }
}

\author{
  Marlene Grieskamp \\
  Interactive Intelligence \\
  University of Washington \\
  Seattle, WA \\
  \texttt{markamp@uw.edu} \\
   \And
  Chaytan Inman \\
  Interactive Intelligence \\
  University of Washington \\
  Seattle, WA \\
  \texttt{chaytan@uw.edu} \\
  \And
  Shaun Lee \\
  Interactive Intelligence \\
  University of Washington \\
  Seattle, WA \\
  \texttt{shauncl8@uw.edu} \\
}

\begin{document}
\maketitle
\begin{abstract}
Cellular automata captivate researchers due to the emergent, complex individualized behavior that simple global rules of interaction enact. Recent advances in the field have combined cellular automata with convolutional neural networks to achieve self-regenerating images. This new branch of cellular automata is called neural cellular automata [1]. The goal of this project is to use the idea of neural cellular automata to grow prediction machines. We place many different convolutional neural networks in a grid. Each conv net cell outputs a prediction of what the next state will be, and minimizes predictive error. Cells received either their neighbors colors and fitness as input. Each cell’s fitness score described how accurate its predictions were. Cells could also move to explore their environment.
\end{abstract}


\section{Introduction and Context}
Overall, we explore the possibility of emergent behaviors in a multi-agent setting with a simple goal of predicting the next state of the game. By giving each cell agency with a convolutional neural network, we were able to explore more complex multi-agent behavior, giving each agent a short term memory in the form of convolutional parameters, and the ability to explore vs exploit their environment through their output movements. The goal was to explore the types of emerging behavior from groups of CNN agents with a selective pressure toward better predictions of the next state.

\textbf{Question}

In an environment where pixel agents are given mechanisms to self-replicate, compete, communicate, and predict, are these channels enough for the emergence of centralized control from distinct agents?

\textbf{Related Work}

This work was heavily inspired by the paper “Growing Neural Cellular Automata” 1, as well as by the original Game of Life by John Conway 2. The ResNet architecture was also borrowed and modified from this tutorial 3. Finally, the idea of a fitness value for the cells comes from the field of genetic algorithms.
The guiding question and following philosophical implications are deeply connected to the works of Deleuze and Simondon among other philosophers and physicists.

\textbf{Assumptions}
\begin{itemize}
    \item Macroscopic wholes can emerge from discretized atomic agents
    \item Intelligent, accurate predictions of the world emerge from resource scarcity, thus a system requiring accurate predictions to grow imposes resource constraints correlated to those which life imposes on cell division and growth.
    \item We can model primordial selves with self replicating simplified agents given arbitrary, nonstationary bounds on themselves in the form of a pixel.
    \item The foundational goal of life is to self replicate, and that self replication does not imply intelligence, but greater intelligence can often improve self replication.
    \item Through coordination of self replicated beings, the self can expand.

\end{itemize}

\section{Methodology}
\begin{enumerate}
    \item Convolutional neural networks are randomly generated on a 100x100 grid.
        \begin{itemize}
        \item A cell\_grid object keeps track of the object positions.
        \item A grid object keeps track of the vectorized form of the cells at the positions.
        \item An intermediate cell grid object keeps track of the possibly overlapping movements of cells.
        \end{itemize}
    \item We begin the first frame by running forward on all conv nets, passing each one its 3x3 neighboring cells.
    \item The outputs are the predicted colors and fitnesses of 3x3 neighbors in the next frame, and a movement represented by five channels.
        \begin{itemize}
        \item The movement property of cells is updated.
        \end{itemize}
    \item The movements of the cells are computed from the output movement channels. This updates the intermediate cell grid.
    \item If cells overlap, their fitnesses are used to determine which one takes precedence, resolving the intermediate cell grid.
    \item The cell\_grid and grid objects are updated based on the resolved intermediate cell grid to get the new frame.
    \item The output predicted colors and fitnesses are compared to the actual colors and fitnesses of the new frame.
    \item The conv nets are updated using backpropagation.
    \item The colors of each conv net in the grid are updated to reflect the changes in network parameters.
    \item The fitnesses of each conv net in the grid are updated to reflect the loss received.
\end{enumerate}

\section{Evaluation}
The results of this experiment are exploratory; there was no desired outcome, but rather a desire to understand the effects of various design choices on the simulated behavior. To understand the effects, we evaluated the loss of the convolutional neural networks using mean squared error between a cell’s predicted next frame of the game and the actual next frame of the game.

We test the following two methods of prediction:
\begin{itemize}
    \item Each cell has partial observability of the 100x100 grid, and also predicts the partial state of the grid (3x3 input and 3x3 output).
    \item Each cell has partial observability of the 100x100 grid, and predicts the full state of the grid (3x3 input and 100x100 output).
\end{itemize}
We run the following simulations for a different number of initial cells in the range [30, 100, 500] for partial observability and plot the cell losses per epoch for randomly picked cells on the grid, as well as compute the average cell loss as a measure of the overall prediction strength of the conv net for different initial starting states.

For each simulation, we perform a grid search over the following hyperparameters:
\begin{itemize}
    \item Learning Rate = [0.1, 0.05, 0.01, 0.005, 0.001]
    \item Momentum = [0.9, 0.95, 0.97, 0.99]
\end{itemize}
\newpage
\section{Results}
For partial observability with partial state prediction, we plot the losses of 3 cells chosen from the grid:

Experiment 1 - 500 initial cells with 30 epochs
\begin{itemize}
    \item Best hyperparameters: learning rate = 0.01, momentum = 0.99
    \item Avg. cell loss: 2017.8651
\end{itemize}
\textbf{Figures 1.1 - 1.3:}

\begin{figure}[H]
    \centering
    \includegraphics[scale=0.3]{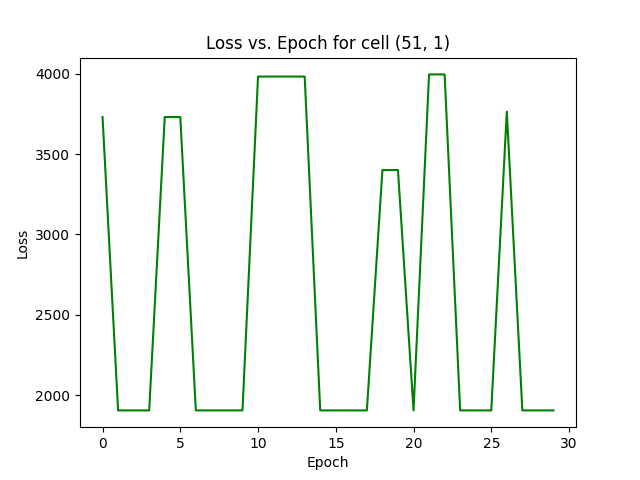}
    \label{fig:my_label}
\end{figure}

\begin{figure}[H]
    \centering
    \includegraphics[scale=0.3]{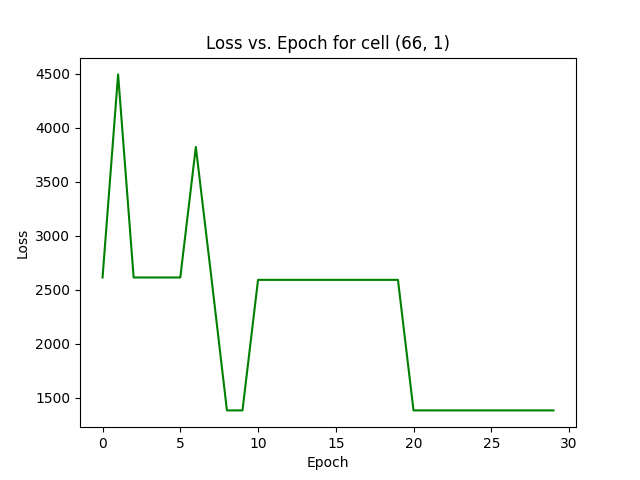}
    \label{fig:my_label}
\end{figure}

\begin{figure}[H]
    \centering
    \includegraphics[scale=0.3]{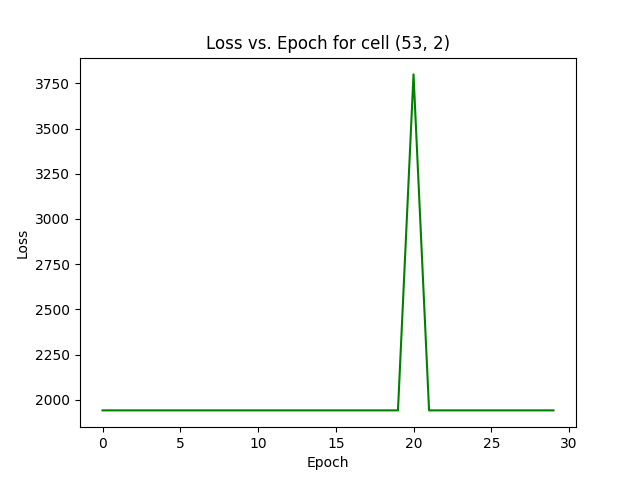}
    \label{fig:my_label}
\end{figure}
\newpage

Experiment 2 - 100 initial cells with 30 epochs
\begin{itemize}
    \item Best hyperparameters:  learning rate = 0.01, momentum = 0.99
    \item Avg. cell loss: 1817.7762
\end{itemize}
\textbf{Figures 2.1-2.3}

\begin{figure}[H]
    \centering
    \includegraphics[scale=0.3]{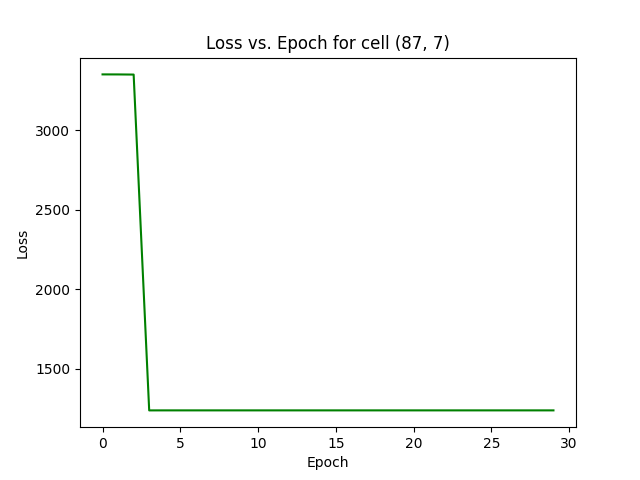}
    \label{fig:my_label}
\end{figure}

\begin{figure}[H]
    \centering
    \includegraphics[scale=0.3]{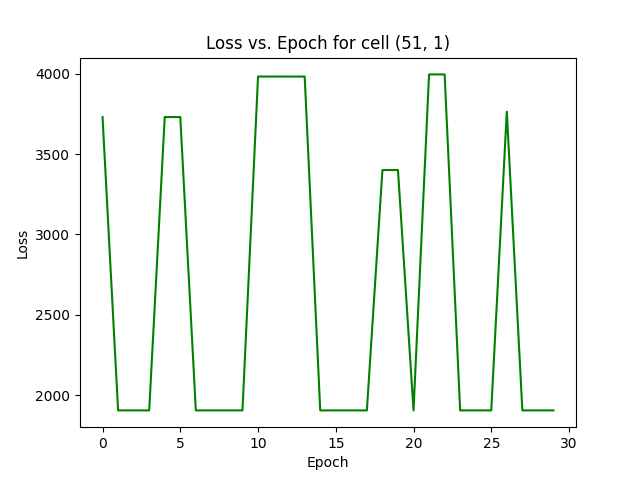}
    \label{fig:my_label}
\end{figure}

\begin{figure}[H]
    \centering
    \includegraphics[scale=0.3]{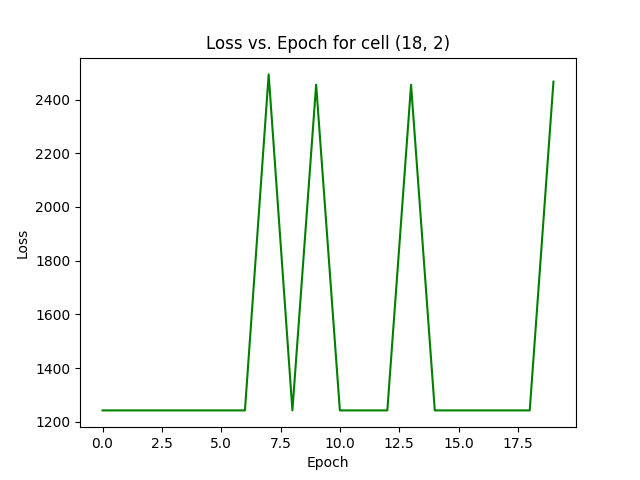}
    \label{fig:my_label}
\end{figure}

\newpage
Experiment 3 - 30 initial cells with 30 epochs
\begin{itemize}
    \item Best hyperparameters: learning rate = 0.005, momentum = 0.97
    \item Avg. cell loss:  1502.6859
\end{itemize}
\textbf{Figures 3.1-3.3}

\begin{figure}[H]
    \centering
    \includegraphics[scale=0.3]{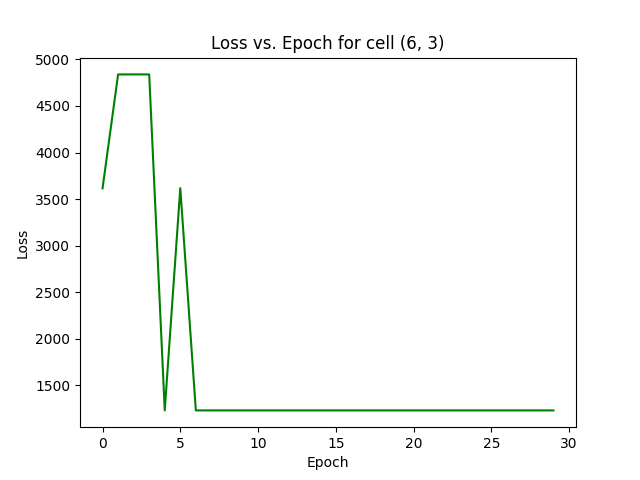}
    \label{fig:my_label}
\end{figure}

\begin{figure}[H]
    \centering
    \includegraphics[scale=0.3]{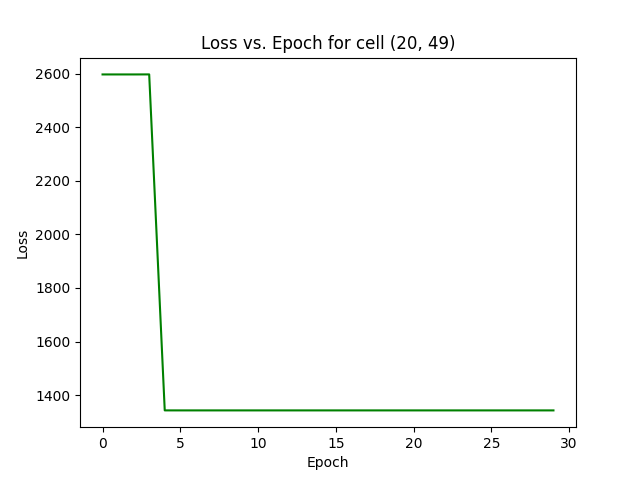}
    \label{fig:my_label}
\end{figure}

\begin{figure}[H]
    \centering
    \includegraphics[scale=0.3]{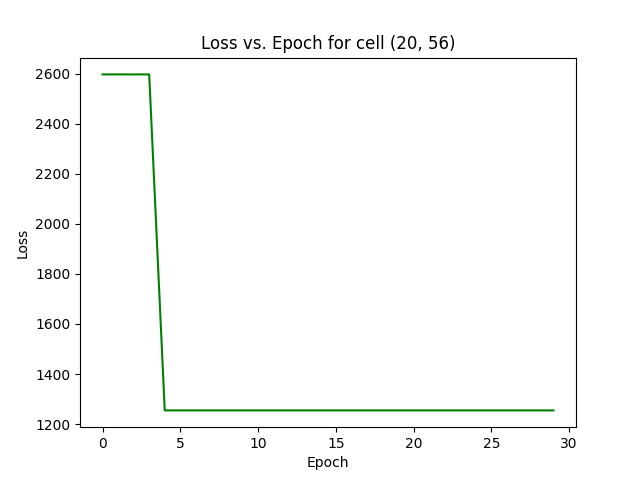}
    \label{fig:my_label}
\end{figure}

\newpage
\section{Discussion}
Using a ResNet architecture, we ran into many issues with upsampling the 3x3 input to a 100x100 output prediction over the whole grid. The result was a very high and unpredictable loss which did not converge. We found that Pytorch’s given methods of upsampling for convolutional neural networks are not well suited for large changes in input sizes, and we needed to add over 20 cells of padding to get the desired output size. We felt this was one reason for the high loss, so we switched to a method of predicting only the cell’s neighbors. This meant that the cell’s input and output were both 3x3x9.

We found that in the former situation where we tried to predict the full state of the grid given partial observability, the loss spiked up and never converged. In the latter experiments 1-3, the loss does spike, but often returns to a low baseline (seen in figures 1.1-1.3, 2.2, 2.3) and may sometimes converge to a lower value (seen in figures 2.1, 3.1-3.3). Although the cause of these results are not exactly certain, we hypothesize that the spiking behavior is due to the cell network learning the position of its neighbors, and then experiencing high loss when neighboring cells move. And for other cells, we hypothesize that the converging behavior occurs when cells move around the grid and encounter few/no neighbors, resulting in more accurate predictions as new cells would not suddenly appear in their field of vision. This may also be supported by the fact that the average cell loss is lower when we start the game with less initial cells (shown in experiments 1-3), as the CNN has a harder time predicting the next state of the game when cells are constantly moving in/out of the 3x3 neighbor grid. We theorize that the loss values are high due to the movement being partially determined by the output of a network and partially determined by stochastic noise (where each cell has a 0.1 probability of taking a random action). This can be difficult for cells to learn and may be one source of high/spiking losses.

The next step may be to train the CNN on the full state/a larger partial state instead of the 3x3 partial state neighbors. By giving the CNN more information, each cell may be better able to predict the next state of the game. Preliminary investigations of this have shown that this significantly increases training time and computation cost, so running this on powerful devices (GPUs/TPUs) is recommended.

\textbf{Future Directions}
Beyond that, we propose the following future implementations:
\begin{itemize}
    \item Weight sharing or mixing between cells
    \begin{itemize}
        \item Instead of each cell having to learn individually, it can instead ‘absorb’ the learned parameters of other cells to improve itself. This may lead to faster convergence and a lower average loss per cell.
    \end{itemize}
    \item Random reproduction of highest fitness cells to allow for clusters of high fitness cells
    \begin{itemize}
        \item Similar to weight sharing: allows cells with the highest predictive power to clone themselves, leading to faster convergence and lower average cell loss.
    \end{itemize}
    \item Better recurrent CNN implementation
    \item Random death of lowest fitness cells
    \begin{itemize}
        \item Removes cells with diverging loss from the game, leading to a more biologically accurate model.
    \end{itemize}
    \item Channels through which convolutional networks can send scalar signals to other networks and in a sense communicate
    \begin{itemize}
        \item Allows for the model to emulate speech/language between cells for more interesting effects.
    \end{itemize}
    \item Neighbor’s predictions of a cell’s fitness impacting the fitness of the cell in question
    \begin{itemize}
        \item Allows for the model to emulate a ‘horde mentality’ between the cells for more interesting effects.
    \end{itemize}

\end{itemize}
\textbf{Possible Use Cases}
\begin{itemize}
    \item Self assembling images
    \item Modeling biological systems – bacteria colonies, multicellular organism formation
    \item Modeling physical systems at a particle level
    \item Economics, and other dynamic multi-agent systems with partial observability
\end{itemize}
\newpage
\section{Demo and Source Code}
The source code for this project is open source:
\href{https://github.com/chaytanc/game-of-intelligent-life/tree/main}{Source Code}

For a deeper dive into the demo as well as philosophical implications derived from the emergence of an individual through cellular automata, see the following:
\href{https://chaytanc.github.io/gil}{Full Demos and Original Publication}

\href{https://giphy.com/gifs/UtGpYehPfGjCkVjyv2}{Full Demo GIF 1}

\href{https://giphy.com/gifs/h1DAaQ4c4RsWDSXPO2}{Full Demo GIF 2}

\begin{figure}[H]
    \centering
    \includegraphics[scale=0.3]{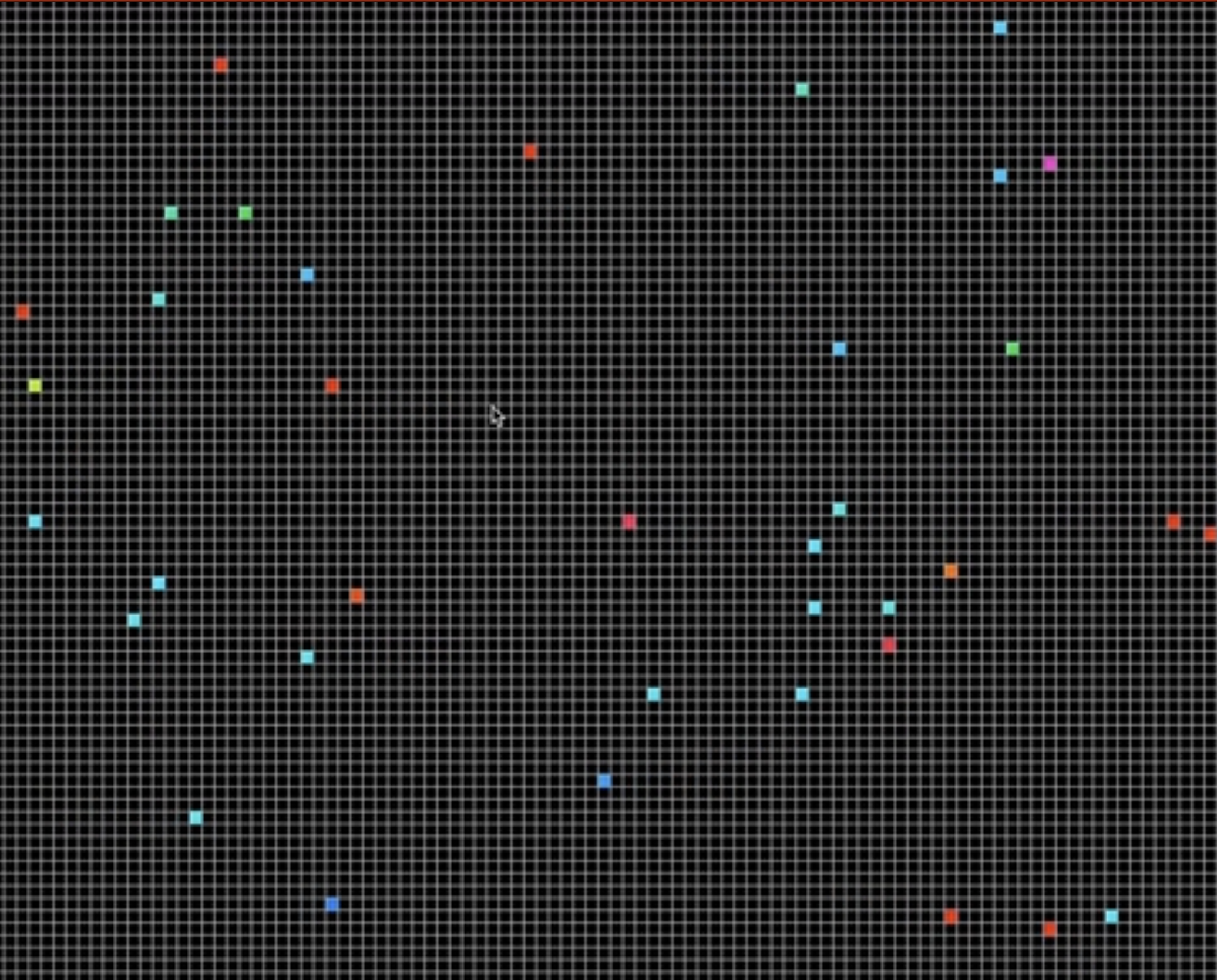}
    \label{fig:my_label}
\end{figure}

This work was created with Interactive Intelligence, a student-led research group at the University of Washington. We would like to thank the I2 advisor, Dr. Eric Chudler, for his support and guidance.

\newpage
\section{References}
\begin{hangparas}{0.25in}{1}
[1] Growing Neural Cellular Automata Mordvintsev, Alexander, et al. “Growing Neural Cellular Automata.” Distill, 27 Aug. 2020, https://distill.pub/2020/growing-ca/.

[2] Conway’s Game of Life “Conway’s Game of Life.” Wikipedia, Wikimedia Foundation, 15 Dec. 2022, https://en.wikipedia.org/wiki/Conway\%27s\_Game\_of\_Life.

[3] \href{https://blog.paperspace.com/writing-resnet-from-scratch-in-pytorch/}{ResNet From Scratch in Pytorch}

[4] Genesis of the Individual Simondon, Gilbert. “Gilbert Simondon - Genesis of the Individual.” Scribd, Scribd, https://www.scribd.com/document/228161394/Gilbert-Simondon-Genesis-of-the-Individual.

[5] Desert Islands Deleuze, Gilles, et al. Desert Islands: And Other Texts 1953-1974. Semiotext(e), 2004.

[6] Relativity v Quantum Mechanics Powell, Corey S. “Relativity v Quantum Mechanics – the Battle for the Universe.” The Guardian, 4 Nov. 2015, https://www.theguardian.com/news/2015/nov/04/relativity-quantum-mechanics-universe-physicists.

[7] General Relativity Two Body Problem “Two-Body Problem in General Relativity.” Wikipedia, Wikimedia Foundation, 29 Oct. 2022, https://en.wikipedia.org/wiki/Two-body\_problem\_in\_general\_relativity.

[8] String Theory Siegel, Ethan. “Why String Theory Is Both a Dream and a Nightmare.” Forbes, Forbes Magazine, 26 Feb. 2020, https://www.forbes.com/sites/startswithabang/2020/02/26/why-string-theory-is-both-a-dream-and-a-nightmare/?sh=38683fba3b1d.

[9] General Relativity “General Relativity.” Wikipedia, Wikimedia Foundation, 14 Dec. 2022, https://en.wikipedia.org/wiki/General\_relativity.

\end{hangparas}

\end{document}